\documentclass[10pt]{article}
\usepackage[textwidth=14cm]{geometry}

\usepackage{natbib}
\usepackage{amsmath,amssymb}
\usepackage{enumitem}
\usepackage{xcolor}
\usepackage{graphicx}
\usepackage{bm}
\usepackage{bbm}
\usepackage{authblk}
\usepackage{hyperref}

\graphicspath{{figs/},{.}}

\newcommand{\until}[1]{\{1,\ldots,#1\}}

\newcommand{\DD}{\mathcal{D}}
\newcommand{\EE}{\mathcal{E}}
\newcommand{\GG}{\mathcal{G}}

\newcommand{\NN}{\mathcal{N}}

\newcommand{\VV}{\mathcal{V}} 
\newcommand{\WW}{W}

\newcommand{\m}{\mathop{\textrm{minimize}}}

\newcommand{\R}{\mathbb{R}}

\newcommand{\CMI}{collective learning}
\newcommand{\CMIspace}{collective learning }
\newcommand{\Upd}{\text{U}}

\newcommand{\ST}{\text{ST}}
\newcommand{\FS}{a}
\newcommand{\eFS}{\tilde{a}}

\newcommand{\lab}{\textsc{lbl}}

\newcommand{\ie}{\emph{i.e.}}
\newcommand{\eg}{\emph{e.g.}}

\begin{document}

\title{Collective Learning}

\author[]{Francesco Farina \\\texttt{franc.farina@unibo.it}}
\affil{Department of Electrical, Electronic and Information Engineering\\
       Alma Mater Studiorum - Universit\`{a} di Bologna\\
       Bologna, Italy
}

\date{}
\maketitle

\begin{abstract}
In this paper, we introduce the concept of collective learning (CL) which exploits the notion of collective intelligence in the field of distributed semi-supervised learning. 
The proposed framework draws inspiration from the learning behavior of human beings, who alternate phases involving collaboration, confrontation and exchange of views with other consisting of studying and learning on their own. 
On this regard, CL comprises two main phases: a self-training phase in which learning is performed on local private (labeled) data only and a collective training phase in which proxy-labels are assigned to shared (unlabeled) data by means of a consensus-based algorithm.
In the considered framework, heterogeneous systems can be connected over the same network, each with different computational capabilities and resources and everyone in the network may take advantage of the cooperation and will eventually reach higher performance with respect to those it can reach on its own.
An extensive experimental campaign on an image classification problem emphasizes the properties of CL by analyzing the performance achieved by the cooperating agents.
\end{abstract}

\section{Introduction}
The notion of \emph{collective intelligence} has been firstly introduced in~\citep{engelbart1962augmenting} and widespread in the sociological field by Pierre L\'{e}vy in~\citep{levy1997collective}. By borrowing the words of L\'{e}vy, collective intelligence ``\emph{is a form of universally distributed intelligence, constantly enhanced, coordinated in real time, and resulting in the effective mobilization of skills}''. Moreover, ``\emph{the basis and goal of collective intelligence is mutual recognition and enrichment of individuals rather than the cult of fetishized or hypostatized communities}''. 

\bigskip

In this paper, we aim to exploit some concepts borrowed from the notion of collective intelligence in a distributed machine learning scenario. In fact, by cooperating with each other, machines may exhibit performance higher than those they can obtain by learning on their own. We call this framework \emph{\CMIspace(CL)}.

Distributed systems\footnote{When talking about distributed systems, the word \emph{distributed} can be used with different meanings. Here, we refer to those networks composed by peer agents, without any central coordinator.} have received a steadily growing attention in the last years and they still deserve a great consideration. In fact, nowadays the organization of computational power and data naturally calls for the distributed systems. It is extremely common to have heterogeneous computing units connected together in possibly time-varying networks. Both computational power and data can be shared or kept private. Moreover, there can be additional globally available resources (such as \emph{cloud}-stored data). Last but not least, preserving the privacy of the local data of the networked systems must be a key requirement. 
This network structure requires the design of tailored distributed algorithms that may let agents benefit from the communication capabilities at disposal, exploit the available computational power and take advantage both of shared and private resources without affecting privacy preservation.

The learning framework we want to address in CL is the one of semi-supervised learning. In particular, we consider problems in which private data at each node are labeled, while shared (and cloud) data are unlabeled. This captures a key challenge in today's learning problems. In fact, while unlabeled data can be easy to retrieve, labeled data are often expensive to obtain (both in terms of time and money) and can be unshareable (due, \eg, to privacy restrictions or trade secrets). Thus, one typically has few local labeled samples and a huge number of (globally available) unlabeled ones. Hybrid problems in which also shared labeled and private unlabeled data are available can be easily included in the proposed framework.

In order to perform CL in the above set-up, we propose an algorithmic idea that is now briefly described. First of all, in order to take advantage of the possible peculiarities and heterogeneity of all the agents in the network, each agent can use a custom architecture for its learning function. The algorithm starts with an initial preliminary phase, called self-training, in which agents independently train their local learning functions on their private labeled data.
Then, the algorithm proceeds with the collective training phase, by iterating through the shared (unlabeled) set. For each unlabeled data, each agent makes a prediction of the corresponding label. Then, by using a weighted average of the predictions of its neighbors (as in consensus-based algorithms), it produces a local proxy-label for the current data and uses such a label to train the local learning function. Weights for the predictions coming from the neighbors are assigned by evaluating the performance on local validation sets.
During the collective training phase, the local labeled dataset are reviewed from time to time in order to give more importance to the local \emph{correctly labeled} data.

We want to emphasize right now that addressing the theoretical properties of the proposed algorithm is beyond the scope of this paper and will be subject to future investigation. Rather, in this work, we present the CL framework for distributed semi-supervised learning, and we provide some experimental results in order to emphasize the features and the potential of the proposed algorithm. 

The paper is organized as follows. The relevant literature for CL is reported in the next section. Then, the problem set-up is formalized and the proposed CL algorithm is presented in details. Finally, an extensive numerical analysis is performed on an image classification problem to evaluate the performance of CL.

\section{Related work}\label{sec:related}
The literature related to this paper can be divided in two main groups: works addressing distributed systems and those involving widely known machine learning techniques that are strictly related to CL.

\medskip

A vast literature has been produced for dealing with distributed systems in different fields, including computer science, control, estimation, cooperative robotics and learning. Many problems arising in these fields can be cast as optimization problems and need to be solved in a distributed fashion via tailored algorithms. Many of them are based on consensus protocols, which allows to reach agreement in multi-agent systems~\citep{olfati2007consensus} and have been widely studied under various network structures and communication protocols~\citep{bullo2009distributed,kar2009distributed,garin2010survey,liu2011consensus,kia2015dynamic}. 
On the optimization side, depending on the nature of the optimization problem to be solved, various distributed algorithms have been developed. Convex problems have been studied within a very large number of frameworks~\citep{boyd2006randomized,nedic2009distributed,zhu2012distributed,ram2010distributed,nedic2010constrained,wei2012distributed,farina2019randomized}, while nonconvex problems have been originally addressed via the distributed stochastic gradient descent~\citep{tsitsiklis1986distributed} and have received recent attention in~\citep{bianchi2013convergence,di2016next,tatarenko2017non,notarnicola2018distributed,farina2019distributed}. In this paper, we consider a different set-up with respect to the one usually found in the above distributed optimization algorithms. In fact, each agent has its own learning function, and hence a local optimization variable that is not related with the ones of other agents. Thus, there is no explicit coupling in the optimization problem. As it will be shown in the next sections, the collective training phase of CL is heavily based on consensus algorithms, but agreement is sought on data and not on decision variables.
Other relevant algorithms and frameworks specifically designed for learning with non-centralized systems include the recent works on distributed learning from constraints~\citep{farina2019LFC}, federated learning~\citep{konevcny2015federated,mcmahan2016communication,konevcny2016federated,smith2017federated} and many other frameworks~\citep{dean2012large,low2012distributed,kraska2013mlbase,li2014scaling,chen2015mxnet,meng2016mllib,chen2016revisiting}. 
Except~\citep{farina2019LFC} and some papers on federated learning, most of these works, however, look for data/model distribution and parallel computation. They usually do not deal with fully distributed systems, because a central server is required to collect and compute the required parameters.

Machine learning techniques related to CL are, mainly, those involving proxy labeling operations on unsupervised data (in semi-supervised learning scenarios). In fact, there exist many techniques in which fictitious labels are associated to unsupervised data, based on the output of one (or more) models that have been previously trained on supervised data only. Co-training~\citep{blum1998combining,nigam2000analyzing,chen2011co} exploit two (or more) views of the data, \ie, different feature sets representing the same data in order to let models produce labeled data for each other. Similarly, in democratic co-learning~\citep{zhou2004democratic} different training algorithms on the same views are exploited, by leveraging off the fact that different learning algorithms have different inductive biases. Labels on unsupervised data are assigned by using the voted majority.
Tri-training~\citep{zhou2005tri} is similar to democratic co-learning, but only three independently trained models are used. In self-training~\citep{mcclosky2006effective,rosenberg2005semi} and pseudo-labeling~\citep{wu2006fuzzy,lee2013pseudo} a single model is first trained on supervised data, then it assigns labels to unsupervised data and uses them for training.
Moreover, strictly related to this work are the concepts of ensemble learning~\citep{tumer1996error,dietterich2000ensemble,wang2003mining,rokach2010ensemble,deng2014ensemble} in which an ensemble of models is used to make better predictions, transfer learning~\citep{bengio2012deep,weiss2016survey} and distillation~\citep{hinton2015distilling} in which models are trained by using other models, and learning with ladder networks~\citep{rasmus2015semi} and noisy labels~\citep{natarajan2013learning,liu2016classification,han2018co}.

\medskip 

To sum up, we point out that this paper utilizes  some of the above concepts both from distributed optimization and machine learning. In particular, we exploit consensus protocols and proxy labeling techniques in order to produce collective intelligence from networked machines.

\section{Problem setup}\label{sec:setup}
In this section the considered problem setup is presented. First, we describe the structure of the network over which agents in the network communicate. Then, the addressed distributed learning setup is described.

\subsection{Communication network structure}
\begin{figure}[h!]
  \centering
  \includegraphics[width=0.8\columnwidth]{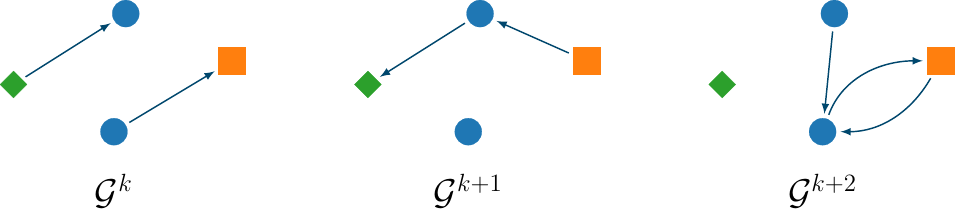}
  \caption{Graphical representation of a time varying graph for which the joint graph $\GG^k\cup\GG^{k+1}\cup\GG^{k+2}$ is strongly connected.}
  \label{fig:diagram-vary}
\end{figure}

We consider a network composed by $N$ agents, which is modeled as a time-varying directed graph $\GG^k=(\VV,\EE^k,\WW^k)$, where $\VV=\until{N}$ is the set of agents, $\EE^k\subseteq\VV\times\VV$ is the set of directed edges connecting the agents at time $k$ and $\WW^k=[w_{ij}^k]$ is the weighted adjacency matrix associated to $\EE^k$. the elements of which satisfy
\begin{enumerate}[label=(\roman*)]
  \item $w_{ii}^k>0$ for all $i\in\VV$,
  \item $w_{ij}^k\geq 0$ if and only if $(j,i)\in\EE^k$,
  \item $\sum_{j=1}^N w_{ij}^k=1$ (\ie, $\WW^k$ is row stochastic),
\end{enumerate}
for all $k=0,1,\dots$. We assume the time-varying graph $\GG^k$ is jointly strongly connected, \ie, there exists $K>0$ such that the graph $\GG^k\cup\GG^{k+1}\cup\dots\cup\GG^{k+K}$ is strongly connected for all $k$ (see Figure~\ref{fig:diagram-vary} for a graphical representation). We denote by $\NN_{i,in}^k$ the set of in-neighbors of node $i$ at iteration $k$ (including node $i$ itself), \ie, $\NN_i^k=\{j\mid (j,i)\in\EE^k\}\cup \{i\}$. Similarly we define the set of out-neighbors of node $i$ at time $k$ as $\NN_{i,out}^k$. The joint-strong connectivity of the graph sequence is a typical assumption and it is needed to guarantee the spread of information among all the agents.

\subsection{Learning setup}
\begin{figure}[h!]
  \centering
  \includegraphics[width=0.4\columnwidth]{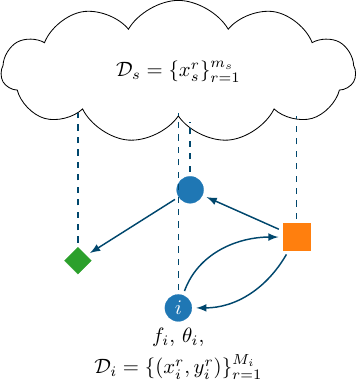}
  \caption{Graphical representation of the considered distributed semi-supervised learning set-up.}
  \label{fig:diagram}
\end{figure}

We consider a semi-supervised learning scenario. Each agent $i$ is equipped with a set of $M_i$ private labeled data points $\DD_i=\{(x_i^r,y_i^r)\}_{r=1}^{M_i}$, where $x_i^r\in\R^d$ is the $r$-th data of node $i$ (with $d$ being the dimension of the input space), and $y_i^r$ the corresponding label. The set $\DD_i$ is divided in a training set and a validation set. The training set consists of the first $m_i<M_i$ samples and is defined as $\DD_{i,T}=\{(x_i^r,y_i^r)\}_{r=1}^{m_i}$, while the validation set is defined as  $\DD_{i,V}=\{(x_i^r,y_i^r)\}_{r=m_i+1}^{M_i}$. Besides, all agents have access to a shared dataset consisting of $m_s$ unlabeled data, $\DD_s=\{x_s^r\}_{r=1}^{m_s}$, with $x_s^r\in\R^d$. 
The goal of each agent is to learn a certain local function $f_i(\theta_i; x)$ (representing a local classifier, regressor, etc.), where we denote by $\theta_i\in\R^{n_i}$ the learnable parameters of $f_i$ and by $x$ a generic input data.
Notice that we are not making any assumption on the local functions $f_i$. In fact, in general, $n_i\neq n_j$ for any $i$ and $j$. A graphical representation of the considered learning setup is given in Figure~\ref{fig:diagram}.

In the experiments, we will consider as a metric to evaluate the actual performance of the agents the accuracy computed on a shared test set $\DD_{test}$. Such a dataset is intended for test purposes only and cannot be used to train the local classifiers.

\section{Collective Learning}\label{sec:collective}
In this section, we present in details our algorithmic idea for CL.
For the sake of exposition, let us consider a problem in which all agents want to learn the same task through their local functions $f_i$. Multi-task problems can be directly addressed in the same way, at the price of a more involved notation.

Each agent in the network is equipped with a local private learning function $f_i$. The structure of each $f_i$ can be arbitrary and different from one agent to the other. For example, $f_1$ can be a shallow neural network with $d$ input units, $f_2$ a deep one with many hidden layers, $f_3$ a CNN and so on. As said, the functions $f_i$ are private for each agent, and, consequently, their learnable parameters $\theta_i$ should not be shared. 

Collective learning consists of two main phases:
\begin{enumerate} 
  \item a preliminary phase (referred to as \emph{self-training}), in which each agent trains its local learning function by using only its private (labeled) data contained in the local training set $\DD_{i,T}$;
  \item a \emph{collective training} phase, in which agents collaborate in order for collective intelligence to emerge.
\end{enumerate} 

\subsection{Self-training}
This first preliminary phase does not require any communication in the network since each agent tries to learn $f_i$ from its private labeled training set $\DD_{t,T}$. 
It allows agents to perform the successive collective training phase after exploiting their private supervised data.

Define $\Phi_i(f_i(\theta_i; x), y)$ as the loss function associated by agent $i$ to a generic datum $(x,y)$. Then, the optimization problem to be addressed by agent $i$ in this phase is
\begin{equation}\label{pb:local}
  \begin{aligned}
    & \m_{\theta_i}
    & & \Psi_i(\theta_i;\DD_{i,T})=\sum_{(x,y)\in\DD_{i,T}} \Phi_i(f_i(\theta_i; x), y).
  \end{aligned}
\end{equation}
Usually, such a problem is solved (meaning that a stationary point is found) by iteratively updating $\theta_i$. Rules for updating $\theta_i$ usually depends on (sub)gradients of $\Psi_i$ or, in stochastic methods, on the ones computed on batches of data from $\DD_{i,T}$.

Consider for simplicity the particular case in which a batch consisting of only one datum is used at each iteration. Most of the current state-of-art algorithms usable in this set-up, starting from the classical SGD~\citep{bottou2010large} to  Adagrad~\citep{duchi2011adaptive}, Adadelta~\citep{zeiler2012adadelta} and Adam~\citep{kingma2014adam}, can be implicitly written as algorithms in which $\theta_i$ is updated by computing
\begin{equation}\label{eq:epoch}
  \theta_i^{h+1} = \Upd_i(\theta_i^h; x_i^h, y_i^h),\quad \text{for } h=1,\dots,m_i,
\end{equation}
where $\theta_i^0$ is some initial condition and $\Upd_i(\theta_i^h; x_i^h, y_i^h)$ denotes the implicit update rule given the current estimate of the parameter $\theta_i^h$ and the data $(x_i^h, y_i^h)\in\DD_{i,T}$ chosen at iteration $h$. 
We leave the update rule implicit, since, depending on the architecture of its own classifier, the available computational power and other factors, each agent can choose the more appropriate way to perform a training step on the current data. As an example, in the classical SGD, the update rule reads $\theta_i^{h+1} = \theta_i^h -\alpha_h \nabla \Phi_i(f_i(\theta_i^h; x_i^h), y_i^h)$ where $\alpha_h$ is a stepsize and $\nabla$ denotes the gradient operator.

In order to approach a stationary point of problem~\eqref{pb:local}, the procedure in~\eqref{eq:epoch} typically needs to be repeated multiple times, \ie, one needs to iterate over the set $\DD_{i,T}$ multiple times. We call~\eqref{eq:epoch} an \emph{epoch} of the training procedure.
Moreover, we denote by $\ST_{i,e}(\hat{\theta}_i)$ the value of $\theta_i$ obtained after a self-training phase started from $\hat{\theta}_i$ and carried out for $e$ epochs.

We assume that the locally available data at each node are relatively few, so that the performance that can be reached on the test set $\DD_{test}$ by solving~\eqref{pb:local} are intrinsically lower than the ones that can be reached by training on a larger and more representative dataset. 

\subsection{Collective training}
This is the main phase of \CMI. It resembles the typical human cooperative behavior that is at the heart of collective intelligence.
Algorithmically speaking, this phase exploits the communication among the agents in the network and uses the shared (unlabeled) data $\DD_s$. 

\subsubsection{Learning from shared data}
In order to learn from shared (unlabeled) data, agents are asked to produce at each iteration proxy-labels for each point in $\DD_s$. 
In general, at each iteration, a batch from the set $\DD_s$ is drawn and processed. To fix the ideas, consider the case in which,, at each iteration $k$, a single sample $x_s^k$ is drawn from the set $\DD_s$. Each node produces a prediction $z_i^k$ for the sample $x_s^k$, by computing $z_i^k = f_i(\theta_i^k; x_s^k)$, and broadcasts it to its out-neighbors $j\in\NN_{i,out}^k$. With the received predictions, each node $i$ produces a proxy-label $\hat{y}_{s,i}^k$ (which we call \emph{local collective label}) for the data $x_s^k$, by converting the weighted average of its own predictions and the ones of its in-neighbors into a label.
Finally, it uses $\hat{y}_{s,i}^k$ as the label associated to $x_s^k$ to update $\theta_i^k$. 
Summarizing, for all $k=1,2,\dots$, each node $i$ draws $x_s^k$ from $\DD_s$ and then it computes
\begin{align}
  z_i^k &= f_i(\theta_i^k; x_s^k)\label{eq:pseudolabel}\\
  \hat{y}_{s,i}^k &= \lab\left[\sum_{j\in\NN_{i,in}^k} w_{ij}^k z_j^k\right]\label{eq:collectivelabel}\\
  \theta_i^{k+1} &=\Upd_i(\theta_i^k; x_s^k, \hat{y}_{s,i}^k )\label{eq:theta}
\end{align}
where we denoted by $\lab[\cdot]$ the operator converting its argument into a label. For example, in a binary classification problem the $\lab$ operator could be a simple thresholding one, \ie, $\lab[x]=0$ if $x<0.5$ and $\lab[x]=1$ if $x\geq 0.5$.

Note that the labeling procedure adopted in this phase highly resembles the human behavior. When unlabeled data are seen, their labels are guessed by resorting to the \emph{opinion} of neighboring agents. 

\subsubsection{Weights computation} 
Let us now elaborate on the choice of the weights $w_{ij}^k$. Clearly, they must account for the expertise and quality of prediction of each agent with respect to the others. In particular, we use as performance index the \emph{accuracy} computed on the local validation sets $\DD_{i,V}$. Let us call $\FS_i^k$ the accuracy obtained at iteration $t$ by agent $i$, and, in order to possibly accentuate the differences between the nodes, let us define 
\begin{equation}\label{eq:acc_weight}
  \eFS_i^k=\exp(\gamma\FS_i^k)
\end{equation} with $\gamma\geq 0$. Then, the weights of the weighted adjacency matrix $\WW^k$ are computed as
\begin{equation}
  w_{ij}^k=
  \begin{cases}
    \frac{\eFS_j^k}{d_i^k}, &\text{if }j\in\NN_{i, in}^k,\\
    0, &\text{otherwise},
  \end{cases}
\end{equation}
where $d_i^k=\sum_{m\in\NN_{i,in}^k} \eFS_m^k$.
By doing so, we guarantee that $\sum_{j=1}^N w_{ij}^k=1$ and weights are assigned proportionally to the performance of each neighboring agent.
Moreover, agents are capable to locally compute the weights to assign to their neighbors, since only locally available information is required.

The value of $\eFS_i^k$ may not be computed at every iteration $k$. In fact, it is very unlikely that it changes too much from one iteration to another. 
Thus, we let agents update their local performance indexes every $T_{i,E}>0$ iterations. In the iterations in which the scores are not updated, they are assumed to be the same as in the previous iteration.

Notice that one can think to different rules for the computation of the weights in the adjacency matrix. For example one can use the F1 score or some other metric in place of the accuracy or assign weights with a different criterion. As a guideline, however, we point out that the weights should always depend on performance of the agents on some (possibly common) task. %
Moreover, the local validation sets should be sufficiently equally informative in order to evaluate agents on an fairly equally difficult task. For example, when available, a common validation set could be used.

\subsubsection{Review step} By taking again inspiration from the human behavior, the collective training phase also includes a review step which is to be performed occasionally by each node (say every $T_{i,R}>0$ iterations for each $i$). Similarly to humans that occasionally review what they have already learned from reliable sources (\eg, books, articles,\dots), agents in the network will review the data in the local set $\DD_{i,T}$ (which are correctly labeled). Formally, every $T_{i,R}$ iterations, node $i$ performs a training epoch on the local data set $\DD_i$, \ie, it modifies step~\eqref{eq:theta} as
\begin{align}
  \hat{\theta}_i^{k+1} &=\Upd_i(\theta_i^k; x_s^k, \hat{y}_{s,i}^k )\\
  \theta_i^{k+1}&=\ST_{i,1}(\hat{\theta}_i^{k+1}).
\end{align}
As it will be shown next, the frequency of the review step plays a crucial role in the learning procedure. A too high frequency tends to produce a sort of overfitting behavior, while too low one makes agent \emph{forget} their reliable data.

\subsection{Remarks}
Before proceeding with the experimental results, a couple of remarks should be done.
The framework presented so far is quite general and can be easily implemented over networks consisting of various heterogeneous systems. In fact, each agent is allowed to use a custom structure for the local function $f_i$. This accounts for, \eg, different systems with different computational capabilities. More powerful units can use more complex models, while those with lower potential will use simpler ones. Clearly, there will be units that will intrinsically perform better with respect to the others, but, at the same time agents starting with low performance (\eg, due to low representative local labeled datasets) will eventually reach higher performance by collaborating with more accurate units.
Finally, CL is intrinsically privacy-preserving since each agent shares with its neighbors only predictions on shared data. Thus, it is not possible to infer anything about the internal architecture or private data of each node, since they are never exposed.

\section{Experimental results}\label{sec:experiments}
Consider an image classification problem in which each agent has a certain number of private labeled images and a huge amount of unlabeled ones is available from some common source (for example the internet).
In this setup, we select the Fashion-MNIST dataset~\citep{xiao2017fashionmnist} to perform an extensive numerical analysis, and CL is implemented in Python by combining TensorFlow~\citep{abadi2016tensorflow} with the distributed optimization features provided by DISROPT~\citep{farina2019disropt}. 
The Fashion-MNIST dataset, consists of $70,000$ 28x28 greyscale images of clothes. Each image is associated with a label from $1$ to $10$, which corresponds to the type of clothes depicted in the image. 
The dataset is divided in a training set $\DD$ with $60,000$ samples and a test set $\DD_{test}$ with $10,000$ samples. 

Next, we first consider a simple communication network and perform a Montecarlo analysis to show the influence of some of the algorithmic and problem-dependent parameters involved in CL.
Then, we compare CL with other non-distributed methods and, finally, an example with a bigger and time-varying network is provided.
The accuracy computed on $\DD_{test}$ is picked as performance metric and the samples in $\DD$ are used to build the local sets $\DD_i$ and the shared set $\DD_s$ in CL.

\subsection{Montecarlo analysis}
\begin{figure}[h!]
  \centering
  \includegraphics[width=0.3\textwidth]{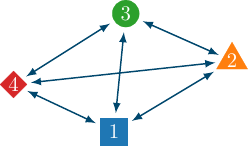}
  \caption{Communication network.}
  \label{fig:graph_complete}
\end{figure}

Consider a simple scenario in which $4$ agents cooperates over a fixed network (represented as a complete graph, depicted in Figure~\ref{fig:graph_complete}) to learn to correctly classify clothes' images.
To mimic heterogeneous agents, the local learning functions $f_i$ of the $4$ agents are as follows.
\begin{enumerate}
  \item $f_1$ is represented as convolutional neural networks (CNN) consisting of (i) a convolutional layer with $32$ filters, kernel size of 3x3 and ReLU activation combined with a maxpool layer with pool size of 2x2;
    (ii) a convolutional layer with $64$ filters, kernel size of 3x3 and ReLU activation combined with a maxpool layer with pool size of 2x2;
    (iii) a convolutional layer with $32$ filters, kernel size of 3x3, ReLU activation and flattened output;
    (iv) a dense layer with 64 units and ReLU activation;
    (v) an output layer with 10 output units and softmax activation.
  \item $f_2$ is represented as a neural network with 2 hidden layers (HL2) consisting of $500$ and $300$ units respectively, with ReLU activation, and an output layer with 10 output units and softmax activation.
  \item  $f_3$ is represented as a neural network with 1 hidden layer (HL1) of $300$ units, with ReLU activation, and an output layer with 10 output units and softmax activation.
  \item $f_4$ is a shallow network (SHL) with 10 output units with softmax activation.
\end{enumerate}

Next, the role of the algorithmic and problem-dependent parameters involved in CL is studied. In particular, we study the performance of the algorithm in terms of the accuracy on the test set by varying: (i) the size of the local training sets, (ii) the review step frequency, and (iii) the parameter $\gamma$ in the weights' computation. %
In all the next simulations, we use $|\DD_{i,V}|=100$ with the samples composing such set randomly picked at each run from the set $\DD$. Moreover, we use the Adam update rule in~\eqref{eq:epoch} and~\eqref{eq:theta}, and a batch size equal to $10$ both in the self-training and in the collective training phases.

\subsubsection{Influence of the local training set size}
\begin{figure}[h!]
  \centering
  \includegraphics[width=0.9\textwidth]{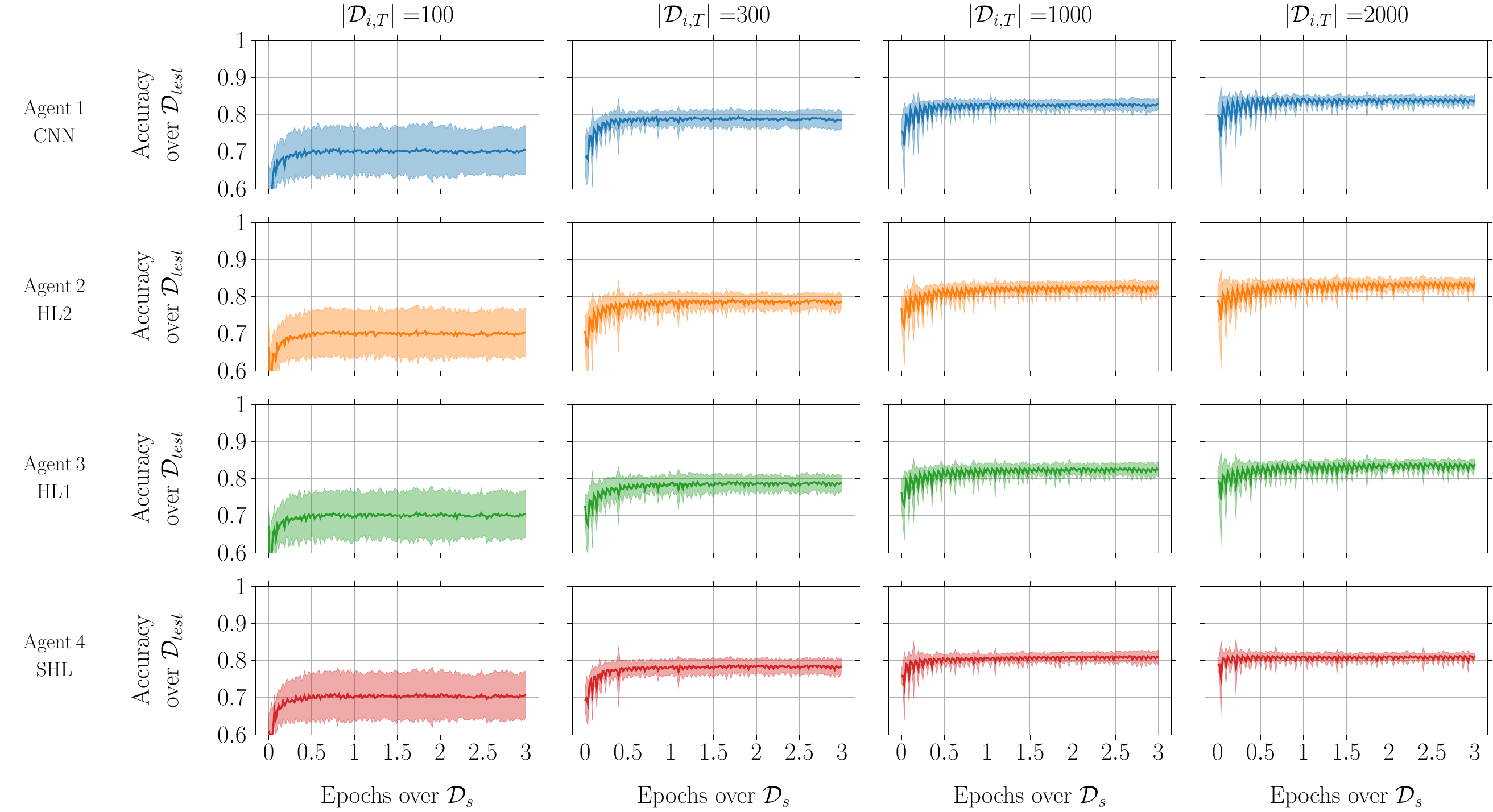}
  \caption{Influence of the local training set size: evolution of the accuracy over $\DD_{test}$ along the algorithm evolution for the $4$ agents, for $\DD_{i,T}\in\{100,300,1000,2000\}$. The solid line represents the average accuracy over $20$ simulations, while the shaded region denotes the average $+/-$ two times the the standard deviation.}
  \label{fig:Montecarlo_samples}
\end{figure}

The number of private labeled samples locally available at each agent is clearly expected to play a crucial role in the performance achieved by each agent. 
In order to show this, we consider $|\DD_{i,T}|\in\{100,300,500,2000\}$. For each value we perform a Montecarlo simulation consisting of $20$ runs. In each run, we randomly pick the samples in each $\DD_{i,T}$ from $\DD$ (along with those in each $\DD_{i,V}$). The remaining samples of $\DD$ are then unlabeled and put in the set $\DD_s$. Then, the algorithm is ran for $3$ epochs over $\DD_s$ with the weights computation and the review step performed every $T_{i,E}=100$ and $T_{i,R}=300$ iterations respectively and $\gamma=100$ in~\eqref{eq:acc_weight}.
The results are depicted in Figure~\ref{fig:Montecarlo_samples}, and two things stand out. First, a higher number of private labeled samples leads to an higher accuracy on the test set $\DD_{test}$. Second, as the number of local samples increases, the variance of the performance tends to decrease. Moreover, it can be seen that all the network architectures reach almost the same accuracy for $|\DD_{i,T}|\in\{100,300\}$, while for $|\DD_{i,T}|\in\{1000,2000\}$, the shallow network is overcome by the other three.

\subsubsection{Influence of the review step frequency}
\begin{figure}[h!]
  \centering
  \includegraphics[width=0.9\textwidth]{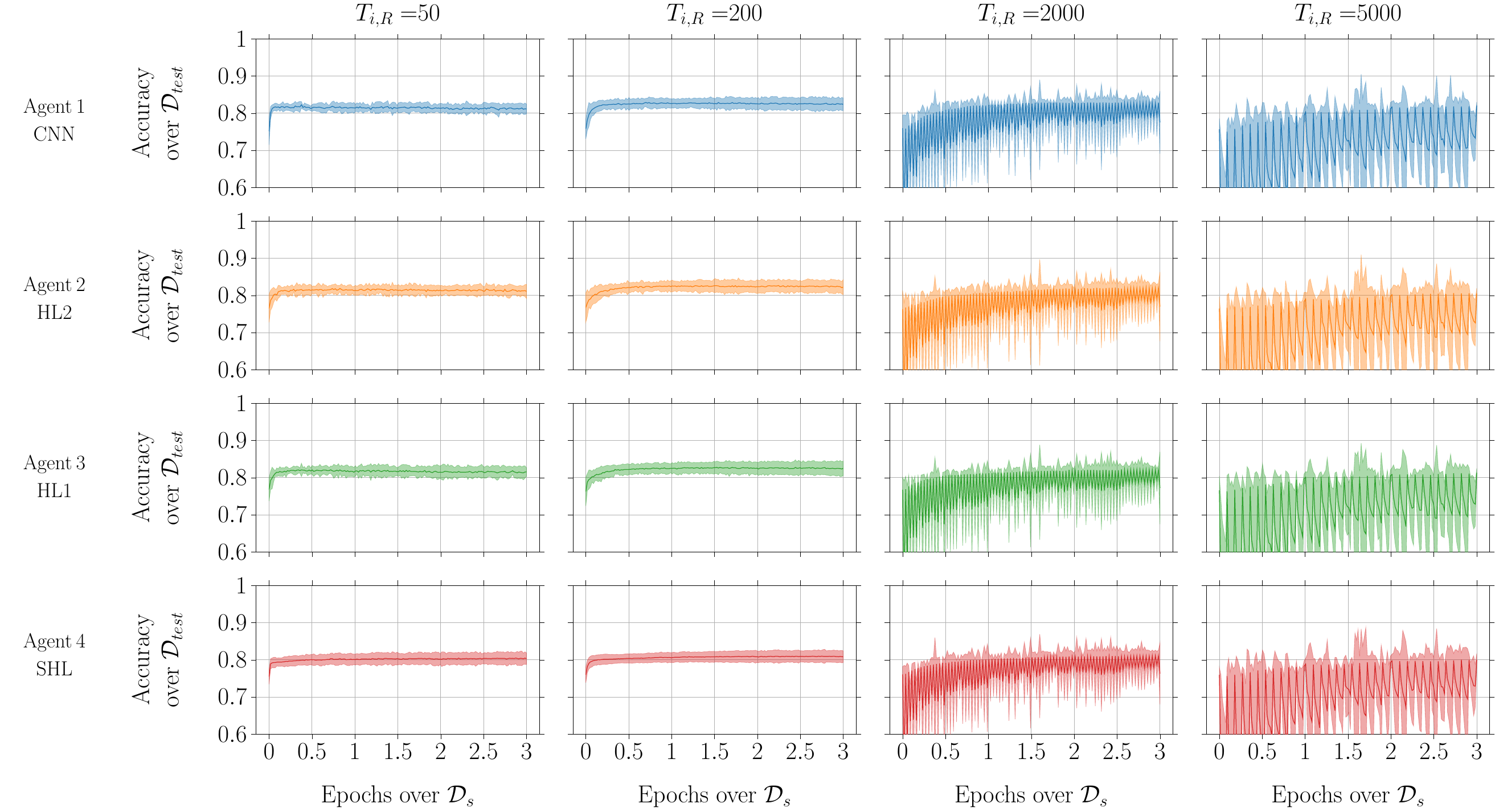}
  \caption{Influence of the review step frequency: evolution of the accuracy computed over $\DD_{test}$ along the collective training phase for the $4$ agents, for $T_{i,R}\in\{50,200,2000,5000\}$. The solid line represents the average accuracy over $20$ simulations, while the shaded region denotes the average $+/-$ two times the the standard deviation.}
  \label{fig:Montecarlo_reviews}
\end{figure}

Also the frequency with which the review step is performed (which is inversely proportional to the magnitude of $T_{i,R}$) influences the performance of the agents. In fact, as humans need to review things from time to time (but not to rarely), also here a too high value for $T_{i,R}$ leads to a performance decay.
We perform a Montecarlo simulation for $T_{i,R}\in\{50, 200, 2000, 5000\}$. For each value we run $20$ instances of the algorithm in which we build each set $\DD_{i,T}$ with $300$ random samples from $\DD$ (along with those in each $\DD_{i,V}$). The remaining samples of $\DD$ are then unlabeled and put in the set $\DD_s$. Then, the algorithm is ran for $3$ epochs over $\DD_s$ with $|\DD_{i,T}|=1000$ for all $i$ and the weights computation performed every $T_{i,E}=100$ with $\gamma=100$ in~\eqref{eq:acc_weight}. The results are reported in Figure~\ref{fig:Montecarlo_reviews}. A higher time interval between two review steps, produce a higher variance and also leads to a lower accuracy. This is due to the fact that if the review step is performed too rarely, agents tends to forget their knowledge on labeled data and start to learn from wrongly labeled samples. Then, when the review occurs they seems to increase again their accuracy. On the contrary, a too high frequency of review step produces a slightly overfitting behavior over the private labeled data. This can be appreciated by comparing in Figure~\ref{fig:Montecarlo_reviews} the cases for $T_{i,R}=50$ and $T_{i,R}=200$. It can be seen that the accuracy on the test set for $T_{i,R}=200$ is higher with respect to the one for $T_{i,R}=50$. A more overfitting behavior can be seen by further reducing $T_{i,R}$.

\subsubsection{Influence of $\gamma$}
\begin{figure}[h!]
  \centering
  \includegraphics[width=\textwidth]{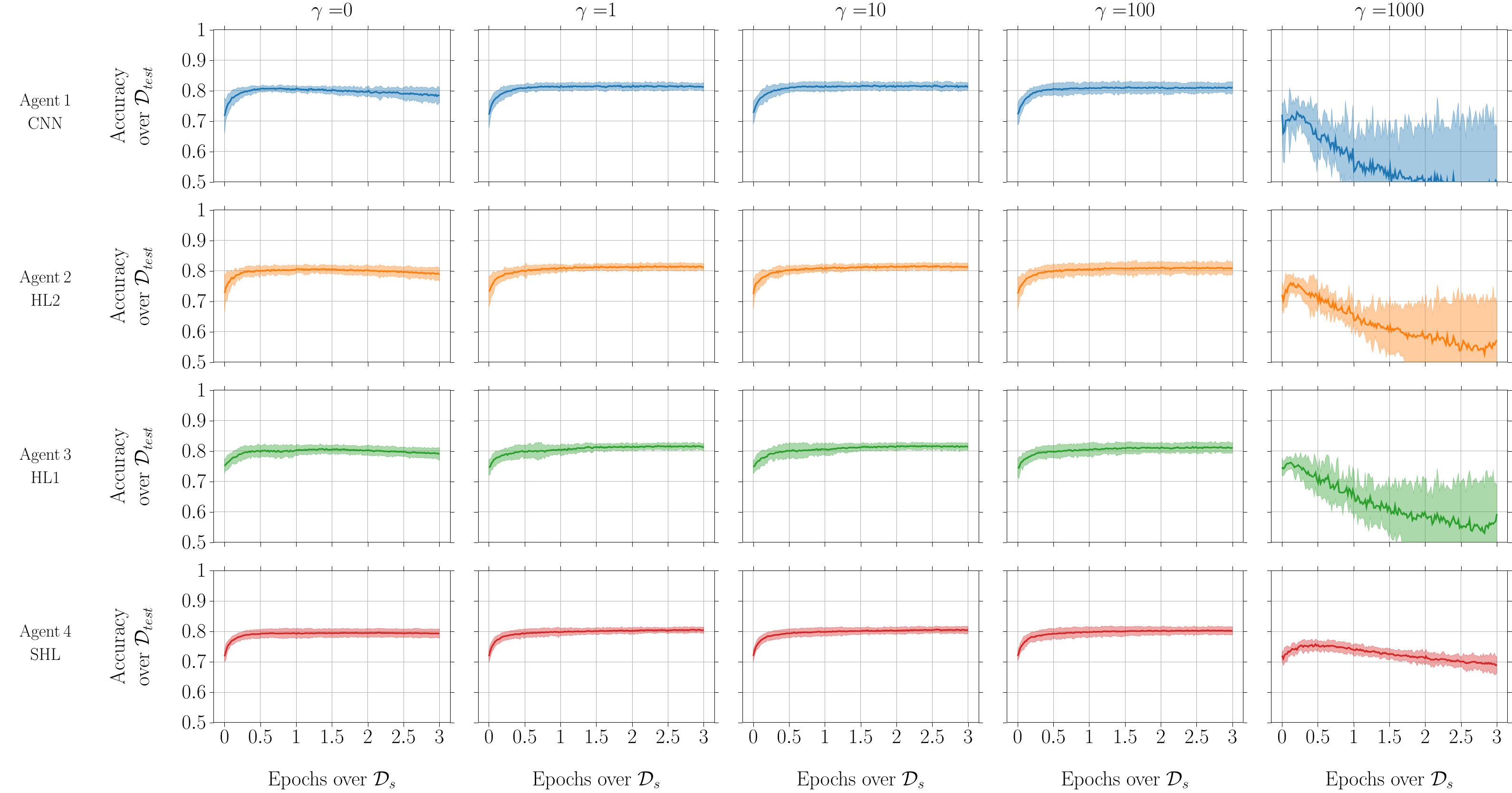}
  \caption{Influence of $\gamma$ (for $|\DD_{i,T}|=500$): evolution of the accuracy computed over $\DD_{test}$ along the collective training phase for the $4$ agents, for $\gamma\in\{0,1,10,100,1000\}$. The solid line represents the average accuracy over $20$ simulations, while the shaded region denotes the average $+/-$ two times the the standard deviation.}
  \label{fig:montecarlo_weights500}
\end{figure}

The last parameter we study is $\gamma$ in \eqref{eq:acc_weight}. A small value of $\gamma$ means that small differences in the local performances results in small differences in the weights. On the contrary, a high value produce a weight near to $1$ for the best agent in the neighborhood. In the extreme case when $\gamma=0$ all agents have in the neighborhood are assigned the same weight, independently of their performance. In Figure~\ref{fig:montecarlo_weights500} the results for a Montecarlo simulation for $\gamma\in\{0,1,10,100,1000\}$ are reported, where we use $|\DD_{i,T}|=500$, $T_{i,R}=500$ and $T_{i,E}=100$.
In this setup, the best accuracy is obtained with $\gamma\in\{1,10,100\}$ with a slightly higher standard deviation for $\gamma=100$. When $\gamma=0$ (\ie, when employing a uniform weighing), the accuracies tend to reach a satisfactory value and, then, start to decrease. This is probably due to the fact that all agents has the same importance and hence, in this case, all of them seems to obtain the performance of the worst of them. Finally, when $\gamma=1000$ there is a substantial performance degradation. This should be caused by the fact that, in the first iterations, there is an agent which is slightly better than the others and leads all the others towards its (wrong) solution.

\begin{figure}[h!]
  \centering
  \includegraphics[width=\textwidth]{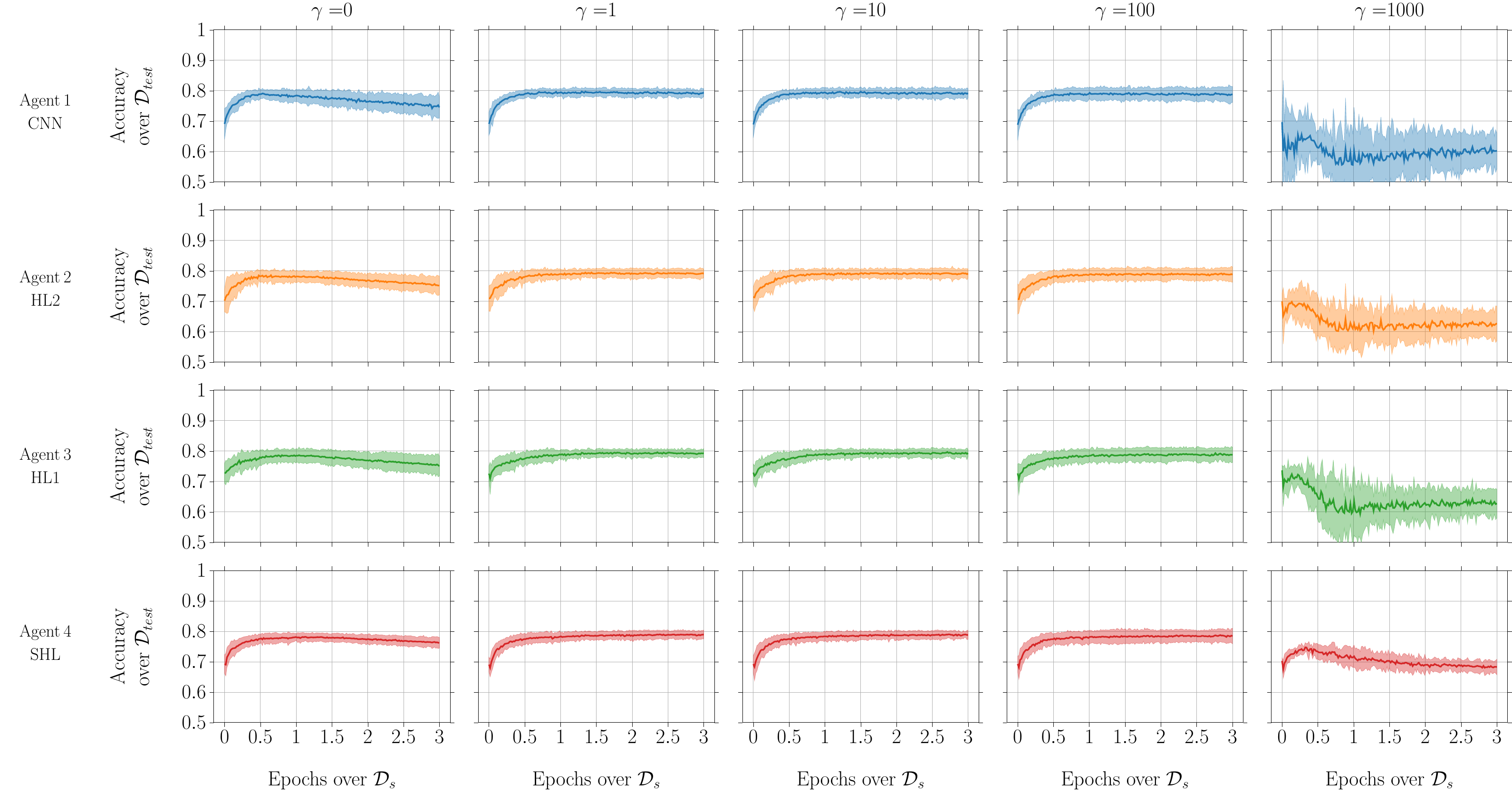}
  \caption{Influence of $\gamma$ (for $|\DD_{i,T}|=300$): evolution of the accuracy computed over $\DD_{test}$ along the collective training phase for the $4$ agents, for $\gamma\in\{0,1,10,100,1000\}$. The solid line represents the average accuracy over $20$ simulations, while the shaded region denotes the average $+/-$ two times the the standard deviation.}
  \label{fig:montecarlo_weights300}
\end{figure}

It is worth mentioning that the influence of $\gamma$ on the performance may vary, depending on the considered setup. 
For example, if we consider $\DD_{i,T}=300$ for all $i$, the results are depicted in Figure~\ref{fig:montecarlo_weights300}. A choice of $\gamma$ in the range $[1,100]$ seems to still work well, while for $\gamma=1000$ a steady state is reached.
Moreover, for bigger generic communication graphs a choice of $\gamma$ too small may not work at all, due to, \eg, having a lot of intrinsically low-performant neighbors whose weight in the creation of the proxy label tends to produce a lot of wrong labels.

\subsection{Comparison with non-cooperative methods}

In this section we compare the results obtained by CL in the presented setup when $|\DD_{i,T}|=500$, $\gamma=100$, $T_{i,E}=100$ and $T_{i,R}=300$ for all $i$ with those obtained by using other (non-cooperative) methods.
In particular, we consider the following two approaches.
\begin{enumerate}
  \item[(ST)] Independently train each learning function over a dataset with the same size of the local private training dataset used in CL, \ie, with $500$ samples. 
  \item[(FS)] Assume that the entire training dataset of Fashion-MNIST is available, and independently train each learning function over the entire dataset $\DD$.
\end{enumerate}
These approaches gives two benchmarks. On one side, the performance obtained by the four learning function with ST coincides with those that can be achieved by the agents by performing the self-training phase only and without cooperating. On the other side, the performance obtained with FS, \ie,~in a fully supervised case, represents the best performance that can be achieved by the selected learning architectures.
In order for CL to be worth for the agents it should lead to better performance with respect to ST and approach as much as possible those obtained by FS.

To compare CL, ST and FS we perform a Montecarlo simulation consisting of 100 runs of each of the three approaches. In each run, the sets $\DD_i$ and $\DD_s$ for CL have been created randomly as in the previous sections, and CL is run for $3$ epochs over $\DD_s$. Similarly, the $500$ samples for training each function of ST are randomly drawn from $\DD$ at each run.
The three approaches are compared in terms of the obtained accuracy on the test set $\DD_{test}$ and the results are reported in Table~\ref{tab:comp500}.

\begin{table}[h!]
  \small
  \centering
    \begin{tabular}{c|c|c|c|c|c|c|}     
                        & \multicolumn{2}{c|}{CL} & \multicolumn{2}{c|}{ST} & \multicolumn{2}{c|}{FS} \\ \hline \hline
    Architecture/agent  & Mean       & Std       & Mean     & Std     & Mean     & Std  \\ \hline
    CNN                 & 0.8149   & 0.0060    & 0.7663  & 0.0127  & 0.9021  & 0.0043   \\ \hline
    HL2                 & 0.8144    & 0.0051    & 0.7670  & 0.0171  & 0.8476  & 0.0058  \\ \hline
    HL1                 & 0.8153    & 0.0049    & 0.7728  & 0.0112  & 0.8465  & 0.0062  \\ \hline
    SHL                 & 0.8065    & 0.0050    & 0.7498  & 0.0077  & 0.8406  & 0.0028  \\ \hline
    \end{tabular}
  \caption{Comparison with non-cooperative methods: mean and standard deviation of the accuracy on the test set for CL, ST and FS.}
  \label{tab:comp500}
\end{table}

It can be seen that CL reaches an higher accuracy (with a lower standard deviation) with respect to ST, thus confirming the benefits obtained through cooperation. The target performance of FS, however are not reached. On this regard we want to point out that the comparison with FS is a bit unfair, since the amount of usable information (in terms of labeled samples) is extremely different. However, it can be shown that with an higher number of samples in $\DD_{i,T}$ an accuracy near to FS can be reached. For example, from Figure~\ref{fig:Montecarlo_samples}, it is clear that with  $\DD_{i,T}=2000$, the learning functions HL2 and HL1 already matches (via CL) the accuracy of FS.

\subsection{Example with a larger, time-varying communication network}
\begin{figure}[h!]
  \centering
  \includegraphics[width=0.7\textwidth]{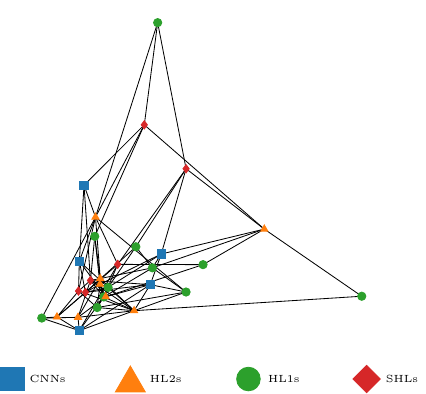}
  \caption{Example with a larger, time-varying communication network: graph example.}
  \label{fig:graph}
\end{figure}

\begin{figure}[h!]
  \centering
  \includegraphics[width=\textwidth]{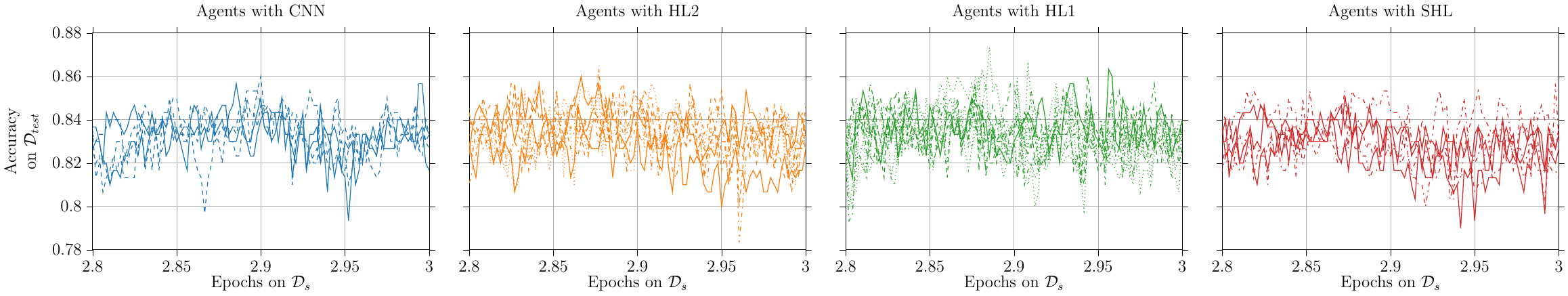}
  \caption{Example with a larger, time-varying communication network: evolution of the accuracy obtained by the various agents over $\DD_{test}$ (in the last part of the third epoch). Each subplot groups together agents equipped with the same learning function.}
  \label{fig:results_large}
\end{figure}

In this section we perform an experiment with a larger network consisting of $30$ agents. Each agent is equipped with a learning function randomly chosen from those introduced in the previous section (CNN, HL2, HL1, SHL). In particular, there are $5$ CNNs, $8$ HL2s, $11$ HL1s and $6$ SHLs. Agents in the network communicate across a time-varying (random) graph that changes every $10$ iterations. Each graph is generated according to an Erd\H{o}s-R\`{e}nyi random model with connectivity parameter $p=0.1$ (see Figure~\ref{fig:graph} for an illustrative example). Each agent is equipped with $|\DD_{i,T}|=300$  training samples randomly picked from the Fashion-MNIST training set. Moreover, we select $T_{i,R}=200$, $T_{i,E}=100$ and $\gamma=10$. We run a simulation in this setup for $3$ epochs over the shared set $\DD_s$ and the results at the end of the simulation are reported in Figure~\ref{fig:results_large} in terms of the accuracy on the test set $\DD_{test}$. It can be seen that all the agents reach an accuracy between $0.8$ and $0.86$. Moreover, in the last iterations, some of them also outperform the target accuracy of FS obtained in the previous section (for HL2, HL1 and SHL). Agents equipped with the CNN, on the other side seems to be unable to reach the accuracy of FS in this setup.

\section{Conclusions}\label{sec:conclusions}
  In this paper we presented the \CMIspace framework to deal with semi-supervised learning problems in a distributed set-up.
  The proposed algorithm allows heterogeneous interconnected agents to cooperate for the purpose of collectively training their local learning functions. The algorithmic idea draws inspiration from the notion of collective intelligence and the related human behavior.
  The obtained experimental results show the potential of the proposed scheme and call for a thorough theoretical analysis of the collective learning framework.

\bibliographystyle{plainnat}
\bibliography{biblio}

\end{document}